\documentclass[letterpaper]{IEEEtran}
\usepackage{latexsym,,amssymb,amsmath,graphicx,epsf,cite,bbm,float}
\usepackage{ifpdf}
\usepackage{epstopdf}

\usepackage{algorithm,algorithmic}
\usepackage{amsmath,amssymb,bm}
\usepackage{amsfonts,dsfont,color,bbm,subcaption}

\usepackage{amsmath,amsthm}
\usepackage{hyperref}

\newtheorem{theorem}{Theorem}

\newtheorem{lemma}[]{Lemma}

\newtheorem{corollary}[]{Corollary}

\newtheorem{remark}[]{Remark}

\newtheorem{definition}[]{Definition}

\newcommand{\be}{\begin{equation}}
\newcommand{\ee}{\end{equation}}
\newcommand{\bea}{\begin{eqnarray}}
\newcommand{\eea}{\end{eqnarray}}

\newcommand{\MB}{\left[\begin{array}}
\newcommand{\ME}{\end{array}\right]}

\newcommand{\ei}{\end{itemize}}
\newcommand{\bi}{\begin{itemize}}

\newcommand{\norm}[1]{\lVert#1\rVert}

\usepackage{amsmath,amsthm}

\DeclareMathOperator*{\argmin}{arg\,min}

\title{Efficient Lipschitzian Global Optimization of Hölder Continuous Multivariate Functions}
\author{\IEEEauthorblockN{Kaan Gokcesu}, \IEEEauthorblockN{Hakan Gokcesu} }
\begin{document}
\maketitle

\begin{abstract}
	This study presents an effective global optimization technique designed for multivariate functions that are Hölder continuous. Unlike traditional methods that construct lower bounding proxy functions, this algorithm employs a predetermined query creation rule that makes it computationally superior. The algorithm's performance is assessed using the average or cumulative regret, which also implies a bound for the simple regret and reflects the overall effectiveness of the approach. The results show that with appropriate parameters the algorithm attains an average regret bound of $O(T^{-\frac{\alpha}{n}})$ for optimizing a Hölder continuous target function with Hölder exponent $\alpha$ in an $n$-dimensional space within a given time horizon $T$. We demonstrate that this bound is minimax optimal.
\end{abstract}

\section{Introduction}

	The challenge of finding the global optimum of a function while minimizing the number of evaluations is a well-known problem in optimization theory. Pinter \cite{pinter1991global} introduced the concept of global optimization, which has become a key technique in many applications, including hyper-parameter tuning in complex learning systems \cite{cesa_book, poor_book}. However, in such systems, the objective function may not exhibit properties that facilitate optimization, such as convexity. As a result, several terms, including derivative-free optimization \cite{rios2013derivative}, black-box optimization \cite{jones1998efficient}, and global optimization \cite{pinter1991global}, have been used to refer to the sequential optimization of unknown and potentially non-convex functions.
	
	The problem of global optimization has garnered significant attention in a variety of research fields, including non-convex optimization \cite{jain2017non, hansen1991number, basso1982iterative}, Bayesian optimization \cite{brochu2010tutorial}, convex optimization \cite{boyd2004convex,nesterov2003introductory,bubeck2015convex}, bandit optimization \cite{munos2014bandits}, and stochastic optimization \cite{shalev2012online, spall2005introduction}. Moreover, global optimization can be readily applied to various real-world applications, such as distribution estimation \cite{gokcesu2018density,willems,coding2}, multi-armed bandits \cite{neyshabouri2018asymptotically,cesa-bianchi,gokcesu2018bandit}, control theory \cite{tnnls3}, signal processing \cite{ozkan}, game theory \cite{tnnls1}, prediction \cite{gokcesu2016prediction,singer}, decision theory \cite{tnnls4}, and anomaly detection \cite{gokcesu2019outlier,gokcesu2016nested,gokcesu2017online}. The importance of global optimization has grown significantly in recent years due to its widespread applicability across different research domains and its potential to address complex problems that are difficult to solve with traditional optimization methods.
	
	Global optimization techniques aim to optimize the objective function $f(\cdot)$ by accessing only its evaluations at specific query points. Several heuristics can be employed to achieve this goal, but we focus on regularity-based approaches \cite{bartlett2019simple,grill2015black}, with a specific emphasis on Lipschitz (and its generalization Hölder) regularity. The Lipschitz regularity approach was first introduced by Piyavskii to optimize univariate Lipschitz continuous functions. The key idea is to construct lower bounding proxy functions for the objective and query their minimum \cite{piyavskii1972algorithm}. Shubert developed a similar method \cite{shubert1972sequential}, which is commonly referred to as the Piyavskii-Shubert algorithm. 
	
	The Piyavskii-Shubert algorithm for Lipschitz continuous functions has been extensively studied and improved upon over the years for various applications \cite{basso1982iterative,schoen1982sequential,mayne1984outer,mladineo1986algorithm,shen1987interval,horst1987convergence,hansen1991number,hansen1995lipschitz,horst2013global}. Over time, the algorithm has been extended to multivariate functions using Taylor expansions \cite{breiman1993deterministic}, and its performance has been accelerated in \cite{baritompa1994accelerations}. Other approaches have also been proposed, such as using smooth auxiliaries \cite{sergeyev1998global} and a variant for differentiable univariate functions presented in \cite{ellaia2012modified}. Brent's method, proposed in \cite{brent2013algorithms}, is suitable for functions defined on a compact interval with a bounded second derivative. Recent work has also focused on univariate global optimization for functions with generalized Lipschitz regularities \cite{gokcesu2021regret}.
	
	Algorithms for global optimization are evaluated based on their convergence to the optimal value of the objective function, which is measured using simple regret \cite{danilin1971estimation}. This metric quantifies the difference between the value of the best query made so far and the value of the optimal solution. For $n$-dimensional Lipschitz continuous objective functions, the Piyavskii-Shubert algorithm has a regret bound of $\tilde{r}_T = O(T^{-\frac{1}{n}})$ \cite{mladineo1986algorithm}. For univariate functions, \cite{hansen1991number} proposed a variant of the Piyavskii-Shubert algorithm that achieves a simple regret of $\epsilon$ within $O(\int_0^1(f(x)-f(x^*)+\epsilon)^{-1}dx)$ queries, which improves upon the previous result of \cite{danilin1971estimation}. Similarly, \cite{ellaia2012modified} improves upon the previous results of \cite{danilin1971estimation,hansen1991number} for univariate functions. LIPO, a variant of the Piyavskii-Shubert algorithm, achieves better simple regret bounds under stronger assumptions \cite{malherbe2017global}. Finally, \cite{bouttier2020regret} investigates the simple regret of the Piyavskii-Shubert algorithm under noisy evaluations.
	
	The Piyavskii-Shubert algorithms have been analyzed for their effectiveness in cumulative regret bounds for univariate functions, which is a more stringent measure than simple regret \cite{gokcesu2021regret}. While these methods work well for Lipschitz continuous functions, the expense of optimizing lower bounding functions to determine queries can be substantial. To address this issue, a recent study \cite{gokcesu2022low} uses pre-determined sampling sets and achieves similar cumulative regret bounds to the Piyavskii-Shubert algorithms for univariate cases. However, these methods are only applicable to objective functions with a single argument. Another work extends this approach to multivariate Lipschitz continuous functions \cite{gokcesu2022efficient}. We extend that approach for use in general Hölder continuous functions. 
	
	In \autoref{sec:pre}, we provide some preliminaries including the problem definition. In \autoref{sec:algorithm}, we provide the algorithm and its implementation. In \autoref{sec:regret}, we provide its cumulative regret analysis and show that minimax optimal regret is achieved with suitable parameter selections.
	
\section{Preliminaries}	\label{sec:pre}
We begin by providing the formal definition of the problem of global optimization in the multivariate case. The objective is to minimize a function $f(\cdot)$ with multiple variables, where $f(\cdot) : \Omega\rightarrow \Re,$ and $\Omega$ is a compact subset of $\Re^n$. In this study, we focus on the case where $\Omega$ corresponds to the unit $n$-dimensional cube, denoted as $\Omega\equiv [0,1]^n$. Despite $f(\cdot)$ not being convex, it is not an arbitrary function and typically exhibits some regularity, such as Holder continuity.

\begin{definition}\label{def:condition}
	Let $f(\cdot)$ satisfy the following regularity:
	\begin{align*}
		|f(x)-f(y)|\leq C\norm{x-y}^{\alpha},
	\end{align*}
	for any $x,y\in\Omega$, where $1>\alpha>0$ is the Hölder exponent, $C>0$ is the regularity constant $\lVert\cdot\rVert$ is the Euclidean norm.
\end{definition}

We adopt an iterative approach to minimize the multivariate function $f(\cdot)$ by iteratively selecting queries based on their evaluations, which are obtained from the previously evaluated queries. The function responsible for selecting the next query point is denoted by $\Gamma(\cdot)$ and is usually derived from Piyavskii-Shubert variants. The aim is to identify the optimal points with the minimum possible number of evaluations. This is equivalent to treating the objective function as a loss function, where each evaluation $f(x_t)$ represents a loss incurred when selecting the query point $x_t\in\Omega$. To analyze our approach's performance, we use the concept of regret, which compares our evaluations with the optimal evaluation. Specifically, we define $x_*$ as the global minimizer of $f(\cdot)$ and use it to compare our selections' evaluations. As $f(\cdot)$ may have arbitrarily high evaluations, we use regret to compare the evaluations of our selections with the optimal evaluation.


There are two types of regret analysis for a time horizon $T$ in global optimization problems. The first is simple regret, which only takes into account the selected points as incurring a loss and not the queried points. The selected point $\tilde{x}_T$ is chosen from the queried points $\{x_t\}_{t=1}^T$ and its evaluation is compared to the minimum possible evaluation of $f(\cdot)$, resulting in the simple regret, $\tilde{r}_T=f(\tilde{x}_T)-\min_{x\in\Omega}f(x)$. The second type of regret analysis is average regret, which considers all queried points as contributing to the loss. The average regret is calculated as the difference between the average evaluation of the queried points and the minimum evaluation of $f(\cdot)$, $$r_T=\frac{1}{T}\sum_{t=1}^{T}f({x}_t)-\min_{x\in\Omega}f(x)$$.

Although evaluating an algorithm's convergence to the optimal evaluation or simple regret can be useful in some problem scenarios, it is may not suitable for global optimization problems due to their inherent difficulty. This is illustrated by the fact that while the basic grid search method's simple regret is minimax optimal, its average regret remains constant \cite{gokcesu2022efficient}. Moreover, a bound on the average regret implies a bound on the simple regret. Hence, we will focus on using average regret for evaluating algorithm performance. Our aim is to develop an algorithm that attains a minimax optimal average regret of $O(T^{-\frac{\alpha}{n}})$.

\section{The Algorithm}\label{sec:algorithm}
In this section, we propose an algorithm for minimizing a non-convex objective function $f(\cdot)$ subject to the constraint defined in \autoref{def:condition}. Although our optimization is performed on an $n$-dimensional unit cube $[0,1]^n$, it can be extended to work with arbitrary cubes of varying sizes and positions through appropriate scaling and translation of inputs. Unlike traditional global optimization algorithms such as Piyavskii-Shubert variants, which use lower bounding proxy functions, our algorithm employs a fixed query construction rule that improves efficiency. The algorithm is outlined as follows:

\begin{enumerate}
	\item Set the parameter $C_0$ as input.
	\item Wrap the domain $\Omega=[0,1]^n$ into a hyper-rectangle $\Theta=[0,\theta^{n-1}]\times[0,\theta^{n-2}]\times\ldots\times[0,\theta^0]$, where $\theta=2^{\frac{1}{n}}$. Extend $f(\cdot)$ analytically by evaluating any $\tilde{x}\in\Theta$ after projecting it onto $\Omega$, i.e., $f(\tilde{x})=f(\argmin_{{x}\in\Omega}\norm{\tilde{x}-{x}})$, which truncates its elements by $1$.
	\item Set the middle point $x_a=\{\theta^{-1},\theta^{-2},\ldots,\theta^{-n}\}$ and its edge vector $v_a=\{\theta^{-1},\theta^{-2},\ldots,\theta^{-n}\}$. Sample $x_a$ and set its evaluation as $f_a=f(\argmin_{x\in\Omega}\norm{x-x_a})$.
	\item Determine two candidate points $x_b$, $x_c$ with their edge vectors $v_b$, $v_c$, and their scores $s_b$, $s_c$, using $x_a$, $v_a$, and $f_a$ as inputs. Add the candidates to the list of potential queries. \label{item:candidate}
	\item Sample the candidate with the lowest score $s'$, remove it from the list, and set it as the query $x'$. Evaluate $f'$ as $f(\argmin_{x\in\Omega}\norm{x-x'})$, and set its edge vector as $v'$. \label{item:sample}
	\item If the number of queries is less than $T$, go to Step \ref{item:candidate} with the inputs $x_a=x'$, $v_a=v'$, $f_a=f'$. Otherwise, stop and return all the queries and their evaluations.
\end{enumerate}

\begin{remark}
	The algorithm has the following characteristics:
	\begin{itemize}
		\item The algorithm maintains the parameters of the potential query points until they are evaluated.
		\item It generates two new potential queries after each query.
		\item The number of potential queries increases linearly with the number of queries made.
		\item Several stopping criteria can be considered based on the specific problem and available computational resources. For instance, the algorithm can terminate after a fixed number of queries or when a certain level of accuracy is achieved.
		\item Each potential query can be represented as a binary string, which is compact, efficient, and allows for easy storage and communication of potential queries between different parts of the algorithm.
	\end{itemize}
\end{remark}

The selection of potential query points in this algorithm differs from that of popular Piyavskii-Shubert variants. Rather than constructing proxy lower bounding functions that pass through the sampled points and minimizing them, the candidate query points are determined by dividing the hyper-rectangle defined by the sampled point $x'$ and its edge vector $v'$ along its largest dimension into two equally sized hyper-rectangles. The center points of these newly created hyper-rectangles serve as the potential new queries. The exact expression for these points is provided below.

\begin{definition}\label{thm:candidate}
	For an objective function $f(\cdot)$, given the queried point $x_a$ and its edge vector $v_a$; the potential queries $x_b$, $x_c$ and their edge vectors $v_b$, $v_c$ are given by
	\begin{align*}
		x_b=&x_a+z, &&v_b=v_a-z,
		\\x_c=&x_a-z, &&v_c=v_a-z,
	\end{align*}
	where $z$ is all- zero except at $I=\argmin_{i\in\{1,2,\ldots,n\}}v_a(i)$, where $v_a(i)$ denotes the $i^{th}$ element of $v_a$. Hence,
	\begin{align*}
		z(I)=0.5{v_a(I)},&&z(i)=0, i\neq I.
	\end{align*}
\end{definition}
\begin{remark}
	The scores $s_b$ and $s_c$ are created as the following
	\begin{align*}
		s_b=&f_a-C_0\norm{v_a}, &&s_c=f_a-C_0\norm{v_a}.
	\end{align*}
\end{remark}
\begin{remark}
	We point out that because of the working structure of the algorithm, if all the scores were offset by some $\epsilon$, the sampled points would not change.
\end{remark}
\begin{lemma}
	For the input $C_0$ and the regularity in \autoref{def:condition} with its respective $C$ and $\alpha$, we have
	\begin{align*}
		\epsilon_0\leq C_0^{\frac{\alpha}{\alpha-1}}C^{\frac{-1}{\alpha-1}},
	\end{align*}
	where $\epsilon_0$ is the minimum $\epsilon$ such that for all possible $x,y\in\Omega$, $\epsilon\geq C\norm{x-y}^\alpha-C_0\norm{x-y}.$
	\begin{proof}
		In terms of the parameters $C_0$, $C$, $\alpha$; an equivalent representation is the following:
		\begin{align}
			&\epsilon_0=\argmin_{\epsilon\in\Re}\epsilon: &&\epsilon\geq C{\triangle}^\alpha-C_0{\triangle}, &&&0\leq\triangle\leq\sqrt{n}
		\end{align}
		To solve for $\epsilon$, we need to investigate the derivative of $C\triangle^\alpha$ with respect to $\triangle$, which is $C\alpha\triangle^{\alpha-1}$. Let $	C_0=C\alpha\triangle_0^{\alpha-1},$
		for some $\triangle_0\in\Re$, where $\triangle_0=\left(\frac{C_0}{C\alpha}\right)^{\frac{1}{\alpha-1}}.$
		Then, we have $\epsilon_0+C_0\triangle_0=C\triangle_0^{\alpha}.$ Hence,
		\begin{align}
			\epsilon_0=&C\left(\frac{C_0}{C\alpha}\right)^{\frac{\alpha}{\alpha-1}}-C_0\left(\frac{C_0}{C\alpha}\right)^{\frac{1}{\alpha-1}}
			\\=&C_0^{\frac{\alpha}{\alpha-1}}C^{\frac{-1}{\alpha-1}}\left(\alpha^{\frac{\alpha}{1-\alpha}}-\alpha^{\frac{1}{1-\alpha}}\right)
			\\\leq& C_0^{\frac{\alpha}{\alpha-1}}C^{\frac{-1}{\alpha-1}},
		\end{align}
	which concludes the proof.
	\end{proof}
\end{lemma}

\begin{lemma}\label{thm:score}
	For an objective function $f(\cdot)$ that satisfies \autoref{def:condition}, when the point $x_a$ is queried with its evaluation $f_a$; with the scores $s_b$ and $s_c$ for the candidate points $x_b$, $x_c$ with their respective edge vectors $v_b$, $v_c$; we have  
	\begin{align*}
		s_b-\epsilon_0\leq&f_a-C\norm{v_a}^\alpha,\\
		s_c-\epsilon_0\leq&f_a-C\norm{v_a}^\alpha,
	\end{align*}
	i.e., the translated scores $s_b-\epsilon_0$ and $s_c-\epsilon_0$ lower bound the evaluation of the respective regions of the queries $x_b$ and $x_c$.
	
	\begin{proof}
		A point $x_b$ with its edge vector $v_b$ covers the hyper-rectangle region $\mathcal{H}_b$ whose center is $x_b$ where the individual distance to the boundary planes are given by its edge vector $v_b$. 
		Because of the regularity in \autoref{def:condition}, we have for $x\in\mathcal{H}_b$; $f(x)\leq f_a-C\norm{x-x_a}^{\alpha}	\leq f_a-C\norm{v_a}^\alpha.$
		Similar arguments follow for $x_c$, $v_c$ and $\mathcal{H}_c$; which concludes the proof.
	\end{proof}
\end{lemma}

In the next section we provide the performance analyses of our algorithm.

\section{Regret analysis}\label{sec:regret}
We start the regret analysis by bounding the regret of a single sampled point.

\begin{lemma}\label{thm:sampleRegret}
	For a given objective function $f(\cdot)$ that satisfies \autoref{def:condition}, let us sample the point $x_b$ with the respective score $s_b$ that was created after the query $x_a$ with its evaluation $f_a$. When the evaluation of the point $x_b$ is $f_b$, we have the following result.
	\begin{align}
		f_b-\min_{x\in\Omega}f(x)\leq C_0\theta\norm{v_b}+C\norm{v_b}^{\alpha}+\epsilon_0,
	\end{align} 
	where $v_b$ is the associated edge vector of the query $x_b$.
	\begin{proof}
		We know that the score of query $x_b$, i.e., $s_b$, completely lower bounds the evaluations in the hyper-rectangle $\mathcal{H}_b$ that is associated with the center point $x_b$ and the edge vector $v_b$, i.e., $s_b-\epsilon_0\leq \min_{x\in\mathcal{H}_b}f(x).$
		By the design of the algorithm, the query list consists of a set of potential queries $\mathcal{X}$ whose associated regions $\mathcal{H}$ are disjoint and their union covers up the whole search domain. Hence, we have $s_b-\epsilon_0\leq \min_{x\in\Omega}f(x).$
	Combining with \autoref{thm:score}, we have
	\begin{align}
		f_a-C_0\norm{v_a}-\epsilon_0\leq \min_{x\in\Omega}f(x).
	\end{align}
	Moreover, because of the regularity from \autoref{def:condition}, we have
	\begin{align}
		f_b\leq f_a+C\norm{v_b}^{\alpha},
	\end{align}
	where the query $x_b$ is created after the sampling of $x_a$ with its evaluation $f_a$. Thus, combining the two results, we get
	\begin{align}
		f_b\leq \min_{x\in\Omega}f(x)+C_0\theta\norm{v_b}+C\norm{v_b}^{\alpha}+\epsilon_0,
	\end{align}
	since $\theta\norm{v_b}=\norm{v_a}$ by design, which concludes the proof.
	\end{proof}
\end{lemma}

This lemma bounds the individual regret of a sampled point in the algorithm with its boundary values and their corresponding functional values, and is a worst case bound



\begin{lemma}\label{thm:regret}
	The algorithm has the following cumulative regret
	\begin{align*}
		\sum_{t=1}^Tf_t-\sum_{t=1}^Tf(x_*)\leq C_0\theta\sum_{t=1}^T\norm{v_t}+C\sum_{t=1}^T\norm{v_t}^{\alpha}+T\epsilon_0.
	\end{align*}
	for the queries points $\{x_t\}_{t=1}^T$, where $\{f_t\}_{t=1}^T$ are their evaluations and $\{v_t\}_{t=1}^T$ are their edge vectors, respectively.
	\begin{proof}
		We run the algorithm for $T$ sampling times. Let this samples be $\{x_t\}_{t=1}^T$ with their respective edge vectors $\{v_t\}_{t=1}^T$ and evaluations $\{f_t\}_{t=1}^T$. Because of Lipschitz continuity of \autoref{def:condition}, for the initial sampling, we have
		\begin{align}
			f_1-f(x_*)\leq C\norm{v_1}^{\alpha}.
		\end{align}
		For the other samplings, i.e., $t\geq2$, we have from \autoref{thm:sampleRegret}
		\begin{align}
			f_t-f(x_*)\leq C_0\theta\norm{v_b}+C\norm{v_b}^{\alpha}+\epsilon_0.
		\end{align}
		Thus, the cumulative regret is bounded by
		\begin{align}
			\sum_{t=1}^Tf_t-\sum_{t=1}^Tf(x_*)\leq C_0\theta\sum_{t=1}^T\norm{v_t}+C\sum_{t=1}^T\norm{v_t}^{\alpha}+T\epsilon_0,
		\end{align}
		which concludes the proof.
	\end{proof}
\end{lemma}		

The cumulative bound is dependent on the sum of powered edge vector norms. To this end, we have the following result.
\begin{lemma}\label{thm:vtnorm}
	We have
	\begin{align*}
		\sum_{t=1}^T\norm{v_t}^\beta\leq&\frac{\theta^{n-\beta}}{\theta^{n-\beta}-1}V^\beta T^{\frac{n-\beta}{n}},
	\end{align*}
	where $V=\norm{v_1}$ for any $\beta$.
	\begin{proof}
		From the construction of the algorithm, with each new sampling, the largest element in the edge vector of the sampled point is halved. Since, at the beginning of the algorithm, the initial edge vector starts as $v_1=\{\theta^{-1},\theta^{-2},\ldots,\theta^{-n}\},$ where $\theta=2^{\frac{1}{n}}$; the norm of the edge vector multiplicatively decreases by $2^{\frac{1}{n}}$. Thus, in the worst-case scenario, we will have
		\begin{align}
			\sum_{t=1}^T\norm{v_t}^\beta\leq\sum_{i=0}^{K-1}2^i\frac{V^\beta}{2^{\frac{i\beta}{n}}}+M\frac{V^\beta}{2^{\frac{K\beta}{n}}},
		\end{align}
		where $2^K-1+M=T$, $2^K\geq M\geq1$ and $V=\norm{v_1}$. Using $M\leq 2^K$, we have
		\begin{align}
			\sum_{t=1}^T\norm{v_t}^\beta\leq&\sum_{i=0}^{K}2^i\frac{V^\beta}{2^{\frac{i\beta}{n}}},
			\\\leq&V^\beta\sum_{i=0}^{K}2^{i\left(\frac{n-\beta}{n}\right)}
			\\\leq&V^\beta\frac{2^{(K+1)\frac{n-\beta}{n}}}{2^{\frac{n-\beta}{n}}-1}
			\\\leq&\frac{\theta^{n-\beta}}{\theta^{n-\beta}-1}V^\beta T^{\frac{n-\beta}{n}},
		\end{align}
	which concludes the proof.
	\end{proof}
\end{lemma}

By using the result of \autoref{thm:vtnorm} for $\beta\in\{\alpha,1\}$, we have the following result.
\begin{theorem}\label{thm:cumreg}
	When $C_0=\lambda_0 CV^{\alpha-1}T^{\frac{1-\alpha}{n}}$ for some $\lambda_0$. We have the following cumulative regret bound
	\begin{align*}
		\sum_{t=1}^Tf_t-\sum_{t=1}^Tf(x_*)\leq O\left((1+\lambda_0+\lambda_0^{\frac{\alpha}{\alpha-1}})CV^\alpha T^{\frac{n-\alpha}{n}}\right)	
	\end{align*}
	\begin{proof}
		From \autoref{thm:regret} and \autoref{thm:vtnorm}, we have
		\begin{align}
			\sum_{t=1}^Tf_t-\sum_{t=1}^Tf(x_*)\leq& C_0\theta\sum_{t=1}^T\norm{v_t}+C\sum_{t=1}^T\norm{v_t}^{\alpha}+T\epsilon_0,
			\\\leq&C_0\frac{\theta^{n}}{\theta^{n-1}-1}V T^{\frac{n-1}{n}}\nonumber
			\\&+C\frac{\theta^{n-\alpha}}{\theta^{n-\alpha}-1}V^\alpha T^{\frac{n-\alpha}{n}}\nonumber
			\\&+C_0^{\frac{\alpha}{\alpha-1}}C^{\frac{-1}{\alpha-1}}T
			\\\leq& O\left((1+\lambda_0+\lambda_0^{\frac{\alpha}{\alpha-1}})CV^\alpha T^{\frac{n-\alpha}{n}}\right),			
		\end{align}
		which concludes the proof.
	\end{proof}
\end{theorem}
Next, we show that when $\lambda_0$ is constant, we have a minimax optimal cumulative regret.
\begin{theorem}
	When $\lambda_0$ is constant, the result in \autoref{thm:cumreg} is minimax optimal.
	\begin{proof}
		Suppose, we have an objective function $f(\cdot)$ whose value is globally optimal at some point $x_*$, i.e., $f(x_*)=\min_{x\in\Omega}f(x)$. Suppose, the function is such that, for some $\epsilon>0$, the $\epsilon$-neighborhood of the global optimum is decreases with the condition in \autoref{def:condition} and constant with $0$ everywhere else, i.e.,
		\begin{align}
			f(x)=\begin{cases}
				C(\norm{x-x_*}^\alpha-C\epsilon^\alpha),& \norm{x-x_*}\leq\epsilon\\
				0,& \norm{x-x_*}>\epsilon
			\end{cases}.
		\end{align}
		Let an algorithm sample the points $\{x_t\}_{t=1}^T$. Its cumulative regret will be given by
		\begin{align}
			R_T= \sum_{t=1}^T(f(x_t)-f(x_*))
		\end{align}
		Suppose, $f(\cdot)$ is such that all the evaluations are valued $0$, i.e., $f(x_t)=0, \forall t$. Hence, $R_T=-Tf(x^*)=CT\epsilon^{\alpha}$. For this, let 
		\begin{align}
			\epsilon_*=\min_\tau\norm{x_\tau-x_*}, 
		\end{align}
		which will maximize $\epsilon$, hence, the regret. 
		Thus, the minimax regret is
		\begin{align}
			R^*_T\geq& CT\min_{\{x_t\}_{t=1}^T}\max_{x_*\in\Omega}\min_\tau\norm{x_\tau-x_*}^{\alpha}\\
			\geq&\Omega(Cn^{\frac{\alpha}{2}}{T^{\frac{n-\alpha}{n}}}),
		\end{align}
		by filling the unit cube with small balls for constant $C$, which concludes the proof.
	\end{proof}
\end{theorem}

\begin{remark}
	For constant $C,\alpha,V$; we can also write the regret in a more compact form as
	\begin{align}
		R_T\leq O\left((\lambda_0+\lambda_0^{-\frac{\alpha}{1-\alpha}})T^{\frac{n-\alpha}{n}}\right),
	\end{align}
	since $\lambda_0+\lambda_0^{-\frac{\alpha}{1-\alpha}}$ is convex and minimum at $\lambda_0=\left(\frac{\alpha}{1-\alpha}\right)^{1-\alpha}$, which lower bounds the expression with $1$.
\end{remark}
	
\begin{corollary}
	When $C_0=T^{\frac{1-\alpha'}{n}}$ for some parameter $\alpha'$, we have
	$R_T\leq O(T^{g(\alpha')})$ for some function $g(\cdot)$. This function is a piecewise linear function that is maximum at its boundaries $g(0)=1$ and $g(1)=1$; and minimum at $g(\alpha)=1-\alpha/n$.
	\begin{proof}
		Let $C_0=T^{\frac{1-\alpha'}{n}}$. Then, 
		\begin{align}
			\lambda_0=A_{\alpha'}T^{\frac{\alpha-\alpha'}{n}},
		\end{align}
		for some constant $A_{\alpha'}$. 
		Hence,
		\begin{align}
			\lambda_0^{-\frac{\alpha}{1-\alpha}}=B_{\alpha'}T^{\frac{\alpha'-\alpha}{1-\alpha}\frac{\alpha}{n}},
		\end{align}
	for some constant $B_{\alpha'}$. 
		The regret will be given by
		\begin{align}
			R_{T}\leq O\left(T^{1-\frac{\alpha'}{n}}+T^{1-\frac{1-\alpha'}{1-\alpha}\frac{\alpha}{n}}\right)
		\end{align}
		Thus, $R_T\leq O(T^{g(\alpha')})$, where
		\begin{align}
			g(\alpha')=\begin{cases}
			{1-\frac{\alpha'}{n}},&0<\alpha'\leq\alpha,
			\\{1-\frac{1-\alpha'}{1-\alpha}\frac{\alpha'}{n}},&\alpha<\alpha'<1,
			\end{cases}
		\end{align}
		which concludes the proof.
	\end{proof}
\end{corollary}
	
\bibliographystyle{ieeetran}
\bibliography{double_bib}
\end{document}